\documentclass{l4dc2024}

\usepackage{array}
\usepackage{pbox}
\usepackage{makecell}
\usepackage{tensor}

\usepackage{array}
\newcolumntype{b}{X}
\newcolumntype{s}{>{\hsize=.4\hsize}X}
\newcolumntype{m}{>{\hsize=.7\hsize}X}
\newcolumntype{t}{>{\hsize=.3\hsize}X}
\newcolumntype{v}{>{\hsize=.2\hsize}X}
\newcolumntype{q}{>{\hsize=.1\hsize}X}

\usepackage{threeparttable}
\title[]{Data-efficient, Explainable and Safe Box Manipulation: Illustrating the Advantages of Physical Priors in Model-Predictive Control}

\usepackage{times}
\coltauthor{\Name{Achkan Salehi} \Email{achkan.salehi@sorbonne-universite.fr}\\
\Name{Stephane Doncieux} \Email{stephane.doncieux@sorbonne-universite.fr}\\
\addr ISIR/CNRS, Sorbonne University, Paris, France}

\begin{document}

\maketitle

\begin{small}
\begin{abstract}%
  Model-based RL/control have gained significant traction in robotics. Yet, these approaches often remain data-inefficient and lack the explainability of hand-engineered solutions. This makes them difficult to debug/integrate in safety-critical settings. However, in many systems, prior knowledge of environment kinematics/dynamics is available. Incorporating such priors can help address the aforementioned problems by reducing problem complexity and the need for exploration, while also facilitating the expression of the decisions taken by the agent in terms of physically meaningful entities. Our aim with this paper is to illustrate and support this point of view via a case-study. We model a payload manipulation problem based on a real robotic system, and show that leveraging prior knowledge about the dynamics of the environment in an MPC framework can lead to improvements in explainability, safety and data-efficiency, leading to satisfying generalization properties with less data.
\end{abstract}


 \section{Introduction}
  \label{sec_intro}

  Model-based Reinforcement Learning (MBRL) \citep{moerland2020model, deisenroth2011pilco} and Model-Predictive Control (MPC) \citep{pinneri2021sample, salehi2022meta, moerland2020model} approaches have increasingly been applied in robotics. These methods display similar adaptivity and generalization to Model Free Reinforcement Learning (MFRL) algorithms but have been shown to be superior in terms of data-efficiency \citep{kaiser2019model,nagabandi2018neural}. However, the data-efficiency of MBRL/MPC relative to Model free approaches does not imply that their are inexpensive. Learning a model of the environment often requires extensive exploration of the state/action space, for which random sampling \textemdash especially in high-dimensional spaces \textemdash is often insufficient \citep{sekar2020planning}, and efficient exploration for both MBRL and MFRL is still an active area of research \citep{sekar2020planning, seo2021state}.

  In addition to this, decisions made by learned policies or estimations made by learned dynamic models are often not readily interpretable by human users without the use of Explainable AI techniques \citep{islam2021explainable}. This results in increased difficulty in debugging and integration in safety-critical systems.

  In many robotic systems however, priors on environment kinematics/dynamics are at least partially available (\textit{e.g.} from classical mechanics). Incorporating such priors into environment models or decision processes can help address the aforementioned problems: it should reduce problem complexity and exploration needs, while facilitating the expression of decisions taken by the agent in terms of physically meaningful concepts. 

  The idea of incorporating knowledge inferred from the rules of physics is not new, and has been studied in gray-box system identification \citep{wileybook, brunton2016sparse} and MBRL with gray-box models \citep{chalup2007machine,lutter2021differentiable}, where it has been shown to present advantages in terms of data-efficiency and safety. Similarly, such ideas have also been explored in works on Physics Informed Neural Networks \citep{sanyal2022ramp}. Our aim with this paper is to support those points of view by providing an illustrative example based on the industrial robot SOTO2 (figure \ref{fig_simulation_description}(a)) which is manufactured by Magazino\footnote{https://www.magazino.eu}, and deployed in production or assembly lines.

  In particular, we focus on controlling the SOTO2's gripper via Model Predictive Control (MPC) with random shooting \citep{nagabandi2018neural, salehi2022meta}. Instead of using a black-box environment model,  we derive a simplified model of the dynamics of the box, and train a neural network to predict its mass distribution using data from a short exploration phase. The predicted mass distribution is then used to compute estimates (\textit{e.g.} moments of inertia) that are necessary for MPC based control using the equations of motion. We evaluate that approach using a pybullet \citep{coumans2021} simulation of the gripper, and show that it has several advantages compared to MPC with a black-box model. Aside from data-efficiency and better explainability, it also enables us to identify potential hazardous cases and halt the operation, which otherwise could lead to failure and even damaged hardware.

\setlength{\fboxsep}{5pt}
\begin{figure}[ht!]
    \centering
    \begin{minipage}[b]{0.4\textwidth}
        \centering
    \fbox{
        \includegraphics[scale=0.24,trim={0cm 300 0cm 0.0cm},clip]{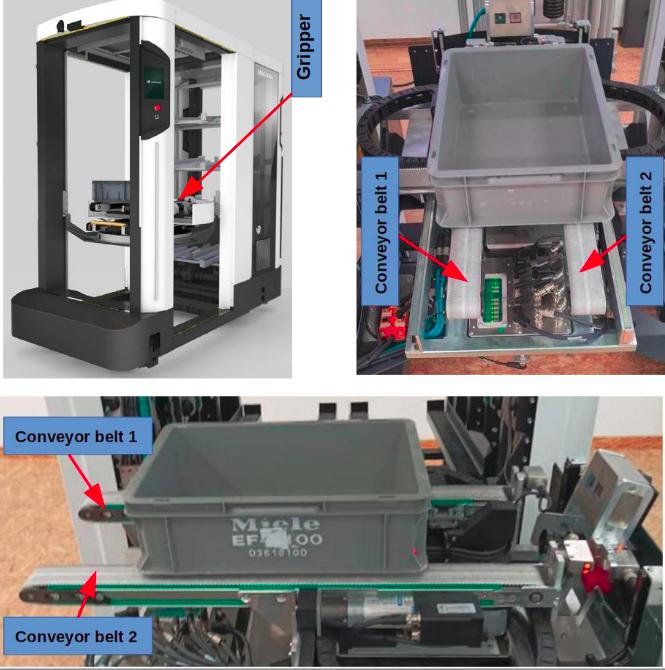}}\\
        (a)
    \end{minipage}
    \begin{minipage}[b]{0.4\textwidth}
        \centering
        \includegraphics[scale=0.24,trim={0cm 0 0cm 0.0cm},clip]{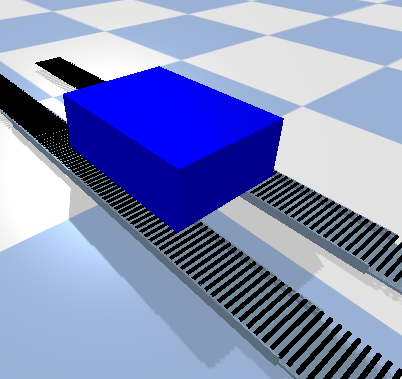}\\
        (b)
    \end{minipage}
    \begin{minipage}[b]{0.26\textwidth}
        \centering
        \includegraphics[scale=0.18,trim={0cm 0 0cm 0.0cm},clip]{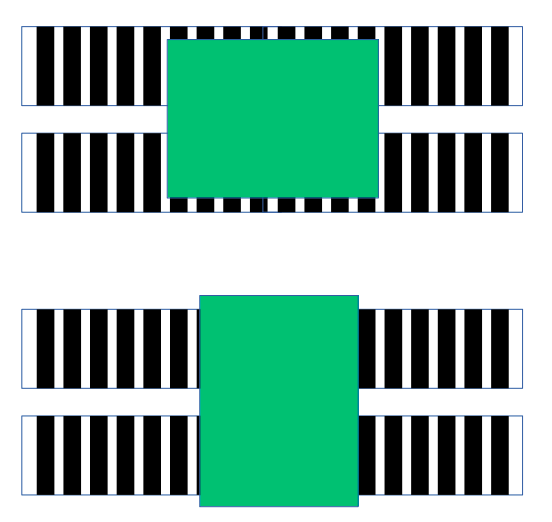}\\
        (c)
    \end{minipage}
    \begin{minipage}[b]{0.26\textwidth}
        \centering
        \includegraphics[scale=0.33,trim={3.4cm 0 3cm 2.37cm},clip]{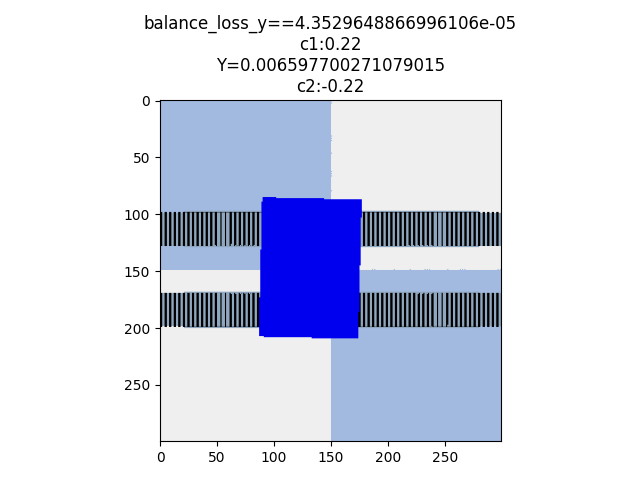}\\
        \centering
        (d)
    \end{minipage}
    \begin{minipage}[b]{0.26\textwidth}
        \centering
        \includegraphics[scale=0.33,trim={3.1cm 0 0cm 2.37cm},clip]{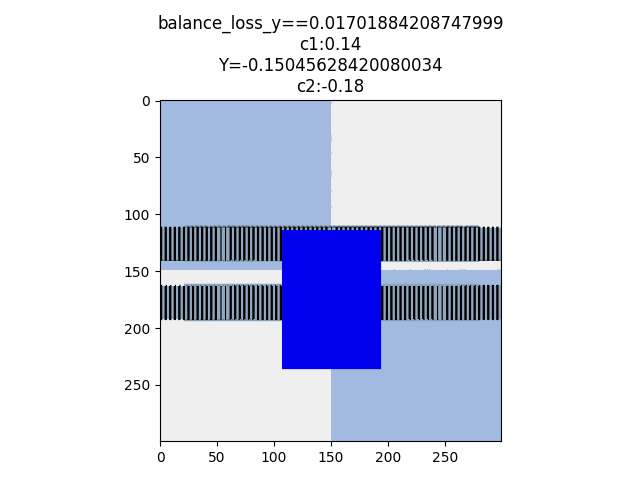}\\
        (e)
    \end{minipage}
    \caption{\small{\textbf{(a)} The SOTO2 robot (left) and its gripper (right), which is composed of two conveyor belts. The motion of each belt cover is set via independent velocity controls, and the distance between the two can be adjusted via position control. The aim of a control episode is to rotate boxes of \textit{varying and unknown} mass distributions by $\frac{\pi}{2}$ in a \textit{safe} manner, \textit{i.e.} the box should either remain well balanced on the conveyors, or manipulation should safely be aborted. \textbf{(b)} Our pybullet simulation of the gripper, based on roller conveyors. \textbf{(c, top)} Initial conditions in each episode. \textbf{(c, bottom)} Target pose. \textbf{(d)} Example final state from a \textit{safe} rotation, where the box stays well balanced. Depending on the mass distribution and the center of mass, this can be difficult to achieve, leading to \textit{unsafe} rotation as shown in \textbf{(e)}.}}
   \label{fig_simulation_description}
\end{figure}

\begin{figure}[h!]
  \centering
  \begin{minipage}[b]{0.2\textwidth}
    \centering
    \includegraphics[width=35mm,trim={0cm 0 0cm 0.0cm},clip]{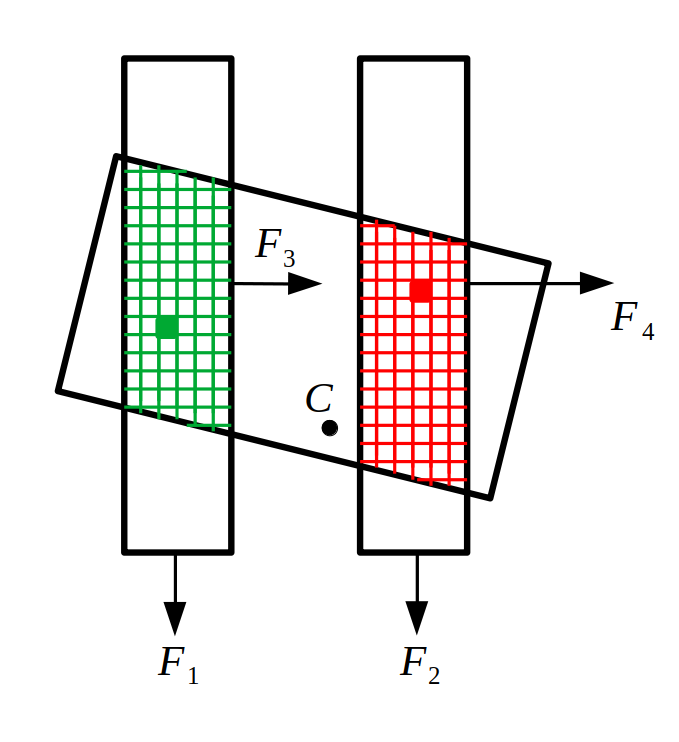}\\
    (a)
  \end{minipage}
  \begin{minipage}[b]{0.2\textwidth}
    \hspace*{0.06cm}
    \centering
     \includegraphics[width=27mm,trim={0cm 0 0cm 0.0cm},clip]{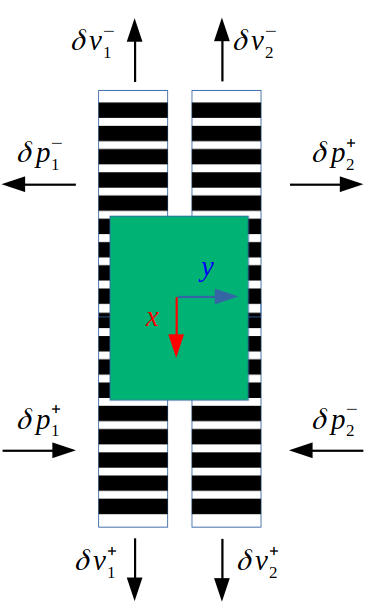}\\
    (b)
  \end{minipage}
  \begin{minipage}[b]{0.2\textwidth}
    \hspace*{0.06cm}
    \centering
    \includegraphics[width=36mm,trim={4cm 1cm 3cm 0.0cm},clip]{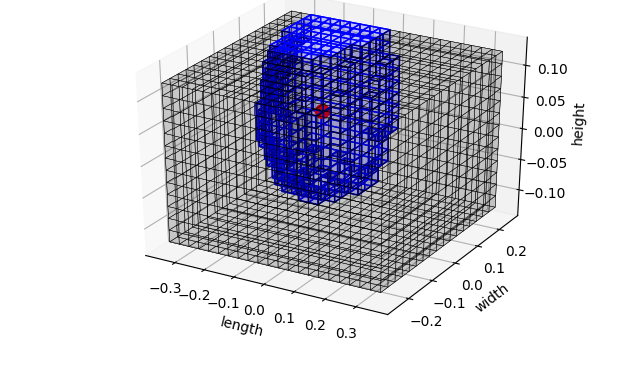}\\
    (c)
  \end{minipage}
  \begin{minipage}[b]{0.2\textwidth}
    \hspace*{0.11cm}
    \centering
    \includegraphics[width=36mm,trim={4.2cm 1cm 2cm 0.0cm},clip]{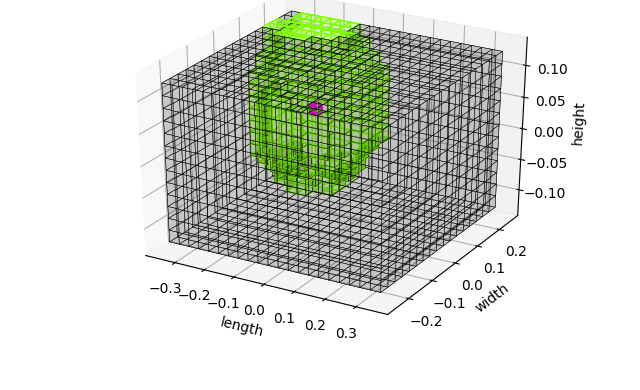}\\
    (d)
  \end{minipage}
   \caption{\small{\textbf{(a)} Top view diagram of the two conveyor belts with illustrative force vectors $F_i$. The shaded volumes, which are those that rest on the conveyor belts, are used to compute torque, friction and the velocity control to force mapping. They are denoted $S_1$ (green) and $S_2$ (red) in the text. \textbf{(b)} Top view illustration of the conveyors with a box on top of them, along with the referential $x,y$ attached to the conveyors. Also shown are the different position and velocity controls. The position controls $\delta p^+_i,\delta p^-_i$ can move the conveyors closer or apart along the $y$ axis, and the velocity controls $\delta v^+_i,\delta v^-_i$ move the surface of the conveyors up/down along the $x$ axis. The action space is thus in $\mathbb{R}^4$. \textbf{(c,d)} Example of ground truth and estimated mass distributions. Each distribution is approximated by a binary occupancy grid and a total mass value $M$, equally distributed between the occupied cells. In our implementation, an infinitesimal positive value is added to the mass of each voxel, in order to avoid degenerate cases with zero friction at contact points. The predicted mass distribution is used to compute the moments of inertia matrix and the center of mass.}}
   \label{fig_dynamics_and_example_distrib_pred}
\end{figure}
  
  \section{Case study: box manipulation using conveyor belts}
  \label{sec_case_study}

We consider the task of \textit{safely} rotating boxes by $\frac{\pi}{2}$ using the SOTO2's gripper, which is composed of two conveyor belts (figure \ref{fig_simulation_description}(a) right), whose individual rolling motions as well as relative distance can be controlled, respectively via velocity and position control (figure \ref{fig_dynamics_and_example_distrib_pred}(b)), resulting in a $4d$ control vector. The robot perceives images from the conveyors and the box via an RGBD camera positioned directly above the gripper, with its axis orthogonal to the ground. Its output is processed by standard computer vision algorithms, from which higher-level observations such as the $3d$ planar pose of the geometric center of the box are available.

  While the dimensions of the box are known in advance, no information is available on box mass distributions, which vary significantly from an episode to the next. The manner in which the box should be rotated depends on the moments of inertia and the center of mass which affects the frictions experienced by the box, and therefore an non-adaptive rotation solution (\textit{e.g.} trivially settings the two conveyor belts to roll in opposite directions) can result in poorly balanced, unsafe rotations, as illustrated in figure \ref{fig_simulation_description}(e). Our aim is to keep the box well balanced during the entire operation, and reach a final state similar to the one from figure \ref{fig_simulation_description}(d).

  We note that safe operation is not possible for all mass distributions. For example, highly skewed mass distributions for which most of the mass falls on one of the conveyor belts are problematic. Such corner-cases constitute an additional motivation for incorporating knowledge about the environment's physics in the control pipeline. Indeed, in the presented problem, estimating the center of mass and moments of inertia can help detect hazardous cases beforehand, which will allow the system to safely abort the manipulation.

  Before describing our solution, we first enunciate two assumptions: \textbf{\textit{(i)}} The mass distribution of the box should not change significantly during an episode. \textbf{\textit{(ii)}} The available computer vision pipeline processing the visual input from the robot should provide $3d$ pose estimation as well as a delimitation of the intersection between the box and the conveyor belts. Both of these tasks can be easily achieved with standard computer vision tools\footnote{Indeed, such pipelines are already available on the robot and have been used in previous works \citep{salehi2022meta}}. In our simulated experiments, we directly get those results from the physics engine.

\subsection{Problem formalization and notations}
  \label{sec_problem_formulation}

We formalize each problem instance (\textit{i.e.} corresponding to a specific mass distribution) as a Multi-Objective Partially Observable Markov Decision Processes (MO-POMDP), that we note $<S,\mathcal{A},\mathcal{F},\{R_i\}_i,\Omega, \Phi,\gamma>$, where $\gamma \in [0,1]$ is a discount factor and $\mathcal{F}:S \times S \times \mathcal{A} \rightarrow [0,1]$ defines the transition probabilities, \textit{i.e.} $\mathcal{F}(s',s,a)=P(s'|s,a)$. The function $\Phi: S \times \Omega \rightarrow [0,1]$ defines the probability $p(o|s)$ of an element $o$ of the observation space $\Omega$ given a state $s$. For this case-study, we define the other components as follows.\\

\noindent\textbf{State space}  Each state $s\in S$ is given by a $5d$ vector which results from concatenating the position of the conveyors along the $y$ axis (see the referential shown in figure \ref{fig_dynamics_and_example_distrib_pred}(b)) to the $3d$ planar pose of the box.\\
\noindent\textbf{Observations}  consist of three components: 1) the 3d pose of the geometric center of the box 2) The surface $S_1$ from the box which intersects the left conveyor, and 3) the surface $S_2$ of the box which intersects the second conveyor. All of these observations can be provided via standard computer vision algorithms on the real robot, and can be directly retrieved in simulations.\\
\noindent\textbf{The action space} is $4d$, with two dimensions corresponding to velocity controls setting the rolling motion of the conveyors, and two dimensions setting the position of the conveyors along the $y$ axis (figure \ref{fig_dynamics_and_example_distrib_pred} (b)).\\
\noindent\textbf{Reward functions}. As our aim is to \textit{safely} rotate the box, we define our set $\{R_i\}$ of objectives as being comprised of two dense reward signals: a pose error reward $R_1$ and a safety criterion $R_2$, which only depend on the state $s'$ that is reached as a result of action $a$:

\begin{equation}
  \begin{split}
    R_1(s',s,a)&=R_1(s')=(||z_{geom}-{z}_{target}||_2^2+\epsilon)^{-1}\\
    R_2(s',s,a)&=R_2(s')=(||r_{geom}-0.5(P_y^1+P_y^2)||_2^2+\epsilon)^{-1}.
  \end{split}
  \label{eq_rewards}
\end{equation}

\noindent with $\epsilon\in\mathbb{R}^{+}$ an infinitesimal value preventing division by zero. In the first equation, $z_{geom}$ is the orientation of the referential attached to the geometric center of the box, and $z_{target}$ is the target orientation written in minimal (\textit{e.g.} Rodrigues) representation. In the second equation, $P_y^1, P_y^2$ respectively denote the $y$ coordinates (figure \ref{fig_dynamics_and_example_distrib_pred}(b)) at which the left conveyor and right conveyor belts are positioned. Intuitively, $R_2$ rewards the control process for centering the geometric center $r_{geom}$ of the box between the conveyor belts. 

In what follows, it will be useful for computing various physical quantities to consider the box as a voxel volume $\Pi$ with resolution $N=N_l \times N_w \times N_h$, effectively considering it as a particle system where the positions of the particles coincide with the center of the voxels. We will note those voxels $\{Q_1, ..., Q_N\}$. For similar reasons, the delimitation of the area of the box resting on the conveyor belts will be important. Those areas, shown as shaded red and green areas in figure \ref{fig_dynamics_and_example_distrib_pred}(a), will respectively be noted $S_1, S_2$. In other terms,

  \begin{equation}
    \begin{split}
      S_1&=\{Q_i | Q_i \in \Pi \text{    is in direct contact with the left conveyor}\}\\
      S_2&=\{Q_i | Q_i \in \Pi \text{    is in direct contact with the right conveyor}\}.\\
    \end{split}
    \label{eq_def_s_i}
  \end{equation}

 Given a mass element in a rigid body (or a particle in a particle system), we will note $r, \dot{r}, \ddot{r}$ respectively its position, velocity and acceleration in world coordinates. When necessary for disambiguation, we will use a subscript, \textit{e.g.} $r_{geom}$ for the position of the geometric center, $r_Q$ for the position of voxel $Q$, etc. We will often use the position, velocity and acceleration $r', \dot{r}', \ddot{r}'$ relative to the center of mass $r_c$, which is defined as

  \begin{equation}
    Mr_c=\int r dm
  \end{equation}

\noindent and can be considered as a particle at which the entire mass of the body $M$ is concentrated, such that its acceleration under the external force $F$ applied to the body obeys the super-particle theorem: $M\ddot{r}_c=F$. We will decompose external forces acting on the box into four component $F_1, ..., F_4$, which each corresponds to one of the actions (figure \ref{fig_dynamics_and_example_distrib_pred} (a,b)). Angular velocity will be defined as $\omega\triangleq \frac{d\phi}{dt}\vec{u}$ with $\vec{u}$ the rotation axis and $\phi$ the angle relative to the center of mass. The moments of inertia tensor relative to the center of mass will be noted $I_c$. Angular momentum will be written $H_c$ and $\tau\triangleq \frac{dH_c}{dt}=I_c\dot{\omega}$ will denote torque. We will also use $\tau=\int r \times df$ to compute the total torque of the rigid body. Finally, the projection of vector $u$ on vector $v$ will be denoted $\textbf{\textrm{proj}}(u,v)\triangleq \frac{v^Tu}{||v||^2}v$.

\subsection{Estimating physical priors from a short exploration phase}
\label{subsec_short_exploration}

As will be discussed in \S\ref{subsec_main_part}, physical quantities such as the center of mass, moments of inertia and the frictions being applied to $S_1$ and $S_2$ have to be estimated. A straight-forward way to enable the estimation of all of these values is to predict the entire mass distribution of the box as a voxel volume, based on an initial short exploration phase. 
In this work, we manually defined a short, \textit{fixed} sequence of controls\footnote{The sequence was comprised of $15$ manually chosen control vectors that apply a small rotation and translation to a box of uniform mass. It was applied for $45$ consecutive timesteps, with each action repeated $3$ times.}, to be sent to the robot at the beginning of each episode. The trajectory from this initial exploration (composed of successive transitions $(s,a,s')$) is fed to a small neural network\footnote{The fully connected network ($\sim 2e6$ parameters) has one head for predicting an occupancy grid to approximate the distribution, and another head that predicts the total mass. The loss is given as a linear combination of a Binary Cross-entropy (BCE) loss computed on the occupancy grid output by the first head, and an MSE loss taking as input the scalar output of the second head. Given the network's size, training and inference are both done on CPU.}, henceforth noted $\mathcal{H}$, which jointly predicts the mass distribution as a 3d occupancy grid and a scalar value as an estimate of the total mass. Figure \ref{fig_dynamics_and_example_distrib_pred} (c,d) show the ground truth and prediction for a random test mass distribution obtained with our network. 

It should be noted that in environments with complex dynamics, learning a black-box model of the environment or learning a Model-Free RL agent requires significant exploration, which is much more costly than the short exploratory phase that the proposed method requires. Another advantage of predicting the mass distribution is that it easily allows us to abort the manipulation early when the detected distribution is physically impossible to safely manipulate.

\subsection{Physical priors based MPC for box rotation}
\label{subsec_main_part}

 We now describe the action evaluation and selection mechanisms after the exploratory phase is over and an estimate of the mass distribution $\hat{\Pi}$ has become available. At each time step $t$, we sample actions $\{a_1^t, ..., a_l^t\}$ via simple random shooting\footnote{Each dimension of the action space is discretized, and at each time step, $N_a$ actions are selected among all the possible vectors formed by this discretizaton. In this work, discretization resulted in $95$ actions, and we set $N_a=50$.}. The total force $F$ applied to the center of mass as a result of an action $a_j^t$ is decomposed into four components $\{F_i\}_{i=1}^4$, such that $F_1, F_3$ correspond to forces applied to $S_1$ and $F_2,F_4$ are applied to $S_2$. Furthermore, $F_1, F_2$ are aligned with the $x$ axis of the referential attached to the conveyor belts and $F_3,F_4$ are aligned with its $y$ axis. An example of such a decomposition is given in figure \ref{fig_dynamics_and_example_distrib_pred}(a). In this section, we assume that the mapping $a_j^t\mapsto \{F_1, F_2, F_3, F_4\}$ between each action and the decomposition of the resulting force is known. The manner in which this mapping is constructed will be discussed in \S\ref{subsec_ctrl_map}.

  To determine the $R_i^j$ associated to an action $a_j^t$, we predict the next state (timestamp $t+\delta t$) that results from applying $a_j^t$ to the system. This can be done in the following way. Let us consider the equations of motion \citep{meriam2020engineering}: we know that for a given mass element \textemdash or voxel \textemdash from the box,

\begin{equation}
  \begin{split}
    \ddot{r}&=\ddot{r}_c + \dot{\omega} \times r' + \omega \times (\omega \times r')\\
    \dot{r}&=\dot{r}_c+\omega \times r'.
  \end{split}
    \label{eq_motion_equation}
\end{equation}

Denoting $F$ the total force that results from action $a_j^t$ on the center of mass, We can write 

\begin{equation}
  \ddot{r}_c(t)\approx \tensor[_0]{\ddot{r}}{_c}(t)+ \tensor[_F]{\ddot{r}}{_c} +\beta
  \label{eq_accel}
\end{equation}

\noindent where $\tensor[_0]{\ddot{r}}{_c}(t)$ is the acceleration of the center of mass without the application of $a_j^t$. Note that the observations produced by the computer vision pipeline include the pose $r_{geom}$ of the geometric center of the box, and that an estimation of its position relative to $r_c$ is available from $\hat{\Pi}$. Therefore, $\tensor[_0]{\ddot{r}}{_c}(t)$ can be approximated in a straight-forward manner from a sliding window of observations. The term $\tensor[_F]{\ddot{r}}{_c}$ is the acceleration that results from applying $a_j^t$ and thus $F$ to the system, and it can easily be recovered from the relation $M\ddot{r}_c=F$. The term $\beta$ is a correction term that will be computed later based on estimated frictions. 

We recover the angular acceleration from the relation

\begin{equation}
\dot{\omega}=I_c^{-1}\tau.
  \label{eq_dotw}
\end{equation}

In order to approximate $\tau$, we leverage the observations $S_1$ and $S_2$ (defined via equation \ref{eq_def_s_i}):

\begin{equation}
  \begin{split}
    \tau &= \sum_{Q\in \hat{\Pi}} r'_{Q} \times \delta F^Q \approx \sum_{Q \in S_1} r'_Q \times \delta F^Q + \sum_{Q \in S_2} r'_Q \times \delta F^Q \\
  &=\sum_{Q \in S_1} r'_Q \times (\delta F^Q_1 + \delta F^Q_3) + \sum_{Q \in S_2} r'_Q \times (\delta F^Q_2+\delta F^Q_4)
  \end{split}
\end{equation}

\noindent where $\delta F_Q$ is the force applied to voxel $Q$, and 

\begin{equation}
  \begin{cases}
    \delta F^Q_i\approx \frac{F_i}{\vert S_1 \vert} & \ \ \ i\in\{1,3\}\\
    \delta F^Q_i\approx \frac{F_i}{\vert S_2 \vert}  & i\in\{2,4\}.\\
  \end{cases}
\end{equation}

Note that the position of the geometric center of the box, $r_{geom}$, is observed via the computer vision pipeline, and since its relative distance from $r_c$ is known from the estimated mass distribution $\hat{\Pi}$, we can compute $r'_Q$ for all voxels $Q$. Similarly, $I_c$ can be trivially computed from $\hat{\Pi}$. Plugging this value and the computed $\tau$ into equation \ref{eq_dotw} allows us to recover $\dot{\omega}$. As for $\omega$ itself, we can approximate it as $\omega\approx \frac{r'_{Q}(t)-r'_{Q}(t-\delta t)}{\delta t}$ using any voxel $Q$, as long as it does not coincide with the center of mass. Similarly, $\dot{r}_c\approx \frac{r_c(t)-r_c(t-\delta t)}{dt}$.

We now have estimates for all the unknowns of equations \ref{eq_motion_equation} and \ref{eq_accel} except for $\beta$. For now, we make a first prediction without accounting for frictions:

\begin{equation}
  \hat{r}^{\textrm{frictionless}}(t+\delta t)=r(t)+\dot{r}(t)\delta t+\frac{1}{2}(\ddot{r(t)}-\beta)\delta t^2
\end{equation}

In order to estimate the frictions, we consider the current and predicted motion of each voxel on the $S_i$. The action $a_j^t$ results in a friction that is in the direction opposite to $\delta \hat{r}(t+\delta t)\triangleq\hat{r}^{frictionless}(t+\delta t)-r(t)$. Note however that the friction that the voxel $Q$ experiences before $a_j^t$ is applied, and which is in the direction opposite to $\dot{r}_Q(t)$, is already accounted for in $\ddot{r}_Q(t)$. Thus, we are only interested in the additional friction component orthogonal to the latter. Noting $\eta_Q^f$ the direction of this friction, we have

\begin{equation}
  \eta_Q^f=-\delta \hat{r}(t+\delta t) - \textbf{\textrm{proj}}\left(-\delta \hat{r}(t+\delta t), -\dot{r}\right)
  \label{eq_fric_1}
\end{equation}

Let us now define the volume $\mathcal{V}_1$ as a superset of $S_1$:

\begin{equation}
    \mathcal{V}_1\triangleq \{Q \in \hat{\Pi} | Q \in S_1 \text{   or   } Q \text{ is above the voxels of  } S_1.\}.
  \label{eq_fric_2}
\end{equation}

In other terms, $\mathcal{V}_1$ is the set of voxels lying on the left conveyor belt, but unlike those included in $S_1$, they are not required to directly touch the conveyor. The volume $\mathcal{V}_2$ is defined analogously. With that notation, the total friction is given by

\begin{equation}
  \begin{split}
    F_f&=\sum_{i\in\{1,2\}}\sum_{Q\in\mathcal{V}(S_i)} \eta_Q^f.\mu_k.g.dM^Q
  \label{eq_fric_3}
  \end{split}
\end{equation}

\noindent where $\mu_k$ is the kinetic friction coefficient, $g\approx 9.8$ and $dM^Q$ is the mass value associated to $Q$. This mass is known from the predicted mass distribution $\hat{\Pi}$. The correction term $\beta$ can now be computed as $\beta=\frac{F_f}{M}$ and plugged into equation \ref{eq_accel}. Similar corrections can be applied to the torque.

The final form of the prediction will then be given by

\begin{equation}
  \hat{r}(t+\delta t)=r(t)+\dot{r}(t)\delta t+\frac{1}{2}\ddot{r(t)}\delta t^2.
\end{equation}

In particular, this allows us to predict $r_{geom}$, which is the only quantity that the safety reward $R_2$ depends on. However, we also need to predict an orientation to compute the pose reward $R_1$. Since we are only interested in the planar motion of the box, orientation is reduced to a single angle $\theta_{geom}$. Analogously to the derivations above, this can be done by considering the Taylor expansion $\hat{\theta}_{geom}(t+\delta t)=\theta_{geom}(t)+\omega \delta t + \frac{1}{2}\dot{\omega}\delta t^2$.\\

\noindent\textbf{Action selection.} The action evaluation process above results in a pair $(R_1(a_j^t), R_2(a_j^t))$ of scores for each of the $l$ sampled actions. The set of potential solutions is given by the Pareto front $A_{front}^t$. However, the solutions in this set are not comparable, and each represent a different compromise between the two $R_i$, that is, between safety and accomplishing the goal. While different selection strategies, ranging from linearly combining the criteria to Choquet integrals \citep{montano2003fuzzy} were considered, we found that randomly selecting an element from $A_{front}^t$ lead to satisfactory results.

  \subsection{Mapping controls to forces}
  \label{subsec_ctrl_map}

 The framework that we have presented assumes that the mapping $a \mapsto \{F_1, F_2, F_3, F_4\}$ is known in advance. As in most commercially available robots, the inner workings of the motion control software applying those given commands to joints is often not exposed to the user or easily accessible, we instead use a simple offline "calibration" phase, which uses data gathered from a few short episodes with a box of uniform mass distribution. Controls are sampled randomly but in a \textit{disjoint} manner, \textit{i.e.} at each timestep, the $4d$ sampled action vector should only have a single non-zero component. This is because our aim is to map actions of the form $a=(a^1, 0, 0 ,0)$, $a=(0, a^2, 0,0)$ (corresponding to velocity controls $\delta v$ in figure \ref{fig_dynamics_and_example_distrib_pred}) respectively to $F_1$ and $F_2$, and actions of the form $a=(0, 0, a^3, 0)$ and $a=(0, 0, 0, a^4)$ (corresponding to the position controls $\delta p$ in figure \ref{fig_dynamics_and_example_distrib_pred}) to $F_3$ and $F_4$, respectively. At the end of each trajectory, the acceleration and velocity of each voxel as well as the friction forces are estimated in a manner similar to that of equations \ref{eq_fric_1}, \ref{eq_fric_2}, \ref{eq_fric_3} above. Assuming that the change of motion at timestamp $t$ is due to the most recently applied action, we can estimate the corresponding force as $F(a)= M\delta \ddot{r}_c - F^1_f - F^2_f$ where $F^1_f$ and $F^2_f$ are the frictions experienced by $S_1$ and $S_2$. Figures \ref{fig_mapping_and_balance} (a, b) show examples of a mapping constructed in this manner using $\sim~400$ transitions for forces $F_1$ and $F_3$. It can be seen that this mapping is indeed uncertain. In this work, we map each action to the mean value of this mapping, and leave the incorporation of uncertainty to the future.

\begin{figure}[ht!]
  \centering
  \begin{minipage}[b]{0.3\textwidth}
  \centering
    \includegraphics[width=37mm,trim={0cm 0 0cm 0.0cm},clip]{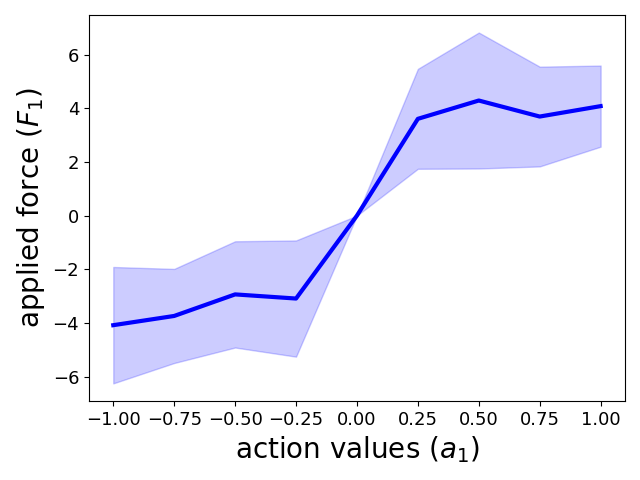}\\
    (a)
  \end{minipage}
  \begin{minipage}[b]{0.3\textwidth}
    \centering
    \includegraphics[width=37mm,trim={0cm 0 0cm 0.0cm},clip]{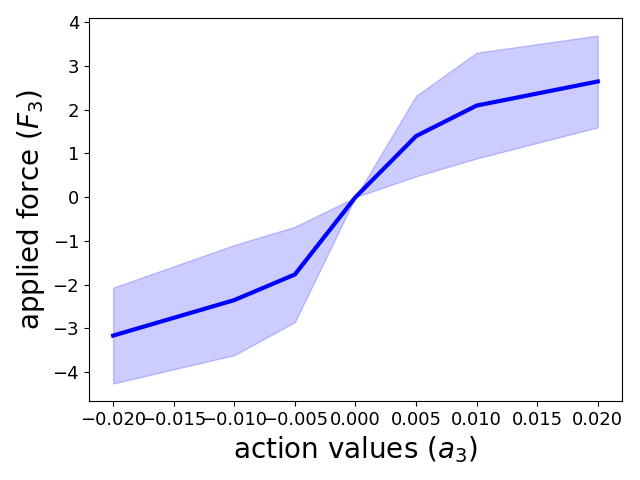}\\
    (b)
  \end{minipage}
  \begin{minipage}[b]{0.3\textwidth}
    \centering
    \includegraphics[width=37mm,trim={0cm 0 0cm 0.0cm},clip]{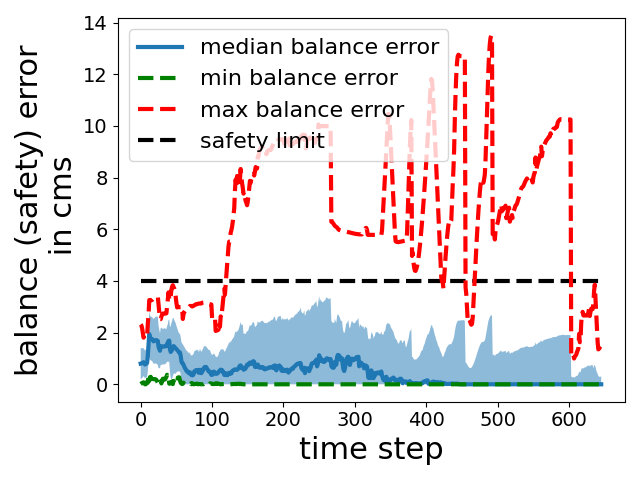}\\
    (c)
  \end{minipage}
  \caption{\small{\textbf{(a,b)} Mappings obtained between applied velocity controls to the different $F_i$. \textbf{(c)} Balance score obtained on the $30$ non-hazardous mass distributions. The shaded area corresponds to one standard deviation around the median. Note that the maximum error (dashed red lines) is caused by an inaccurate prediction from $\hat{\Pi}$.}}
   \label{fig_mapping_and_balance}
\end{figure}
\begin{figure*}[ht!]
  \centering
  %trim={left,lower,right,up}
  \begin{minipage}[b]{0.2\textwidth}
    \centering
    \includegraphics[width=36mm,trim={2cm 1cm 1.5cm 2.0cm},clip]{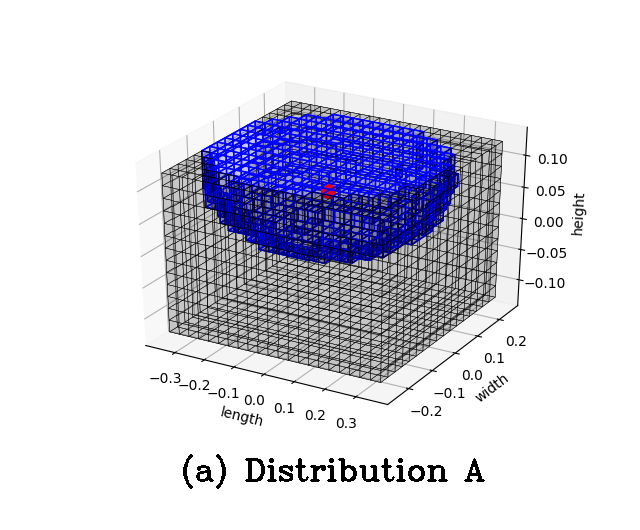}
  \end{minipage}
  \begin{minipage}[b]{0.2\textwidth}
    \centering
    \includegraphics[width=36mm,trim={2cm 1cm 1.5cm 2.0cm},clip]{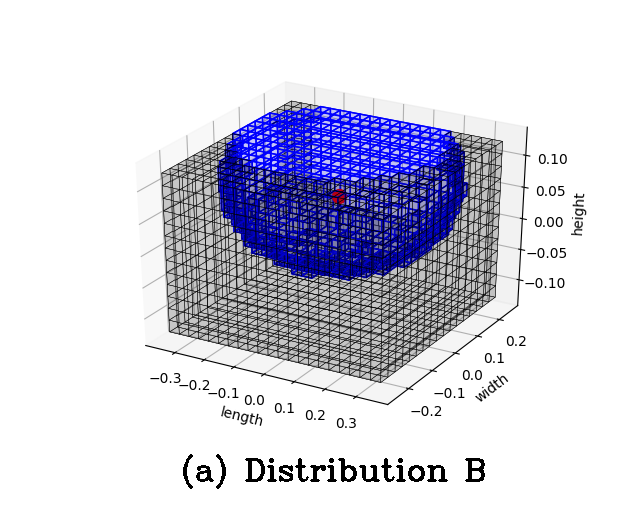}
  \end{minipage}
  \begin{minipage}[b]{0.2\textwidth}
    \centering
    \includegraphics[width=36mm,trim={2cm 1cm 1.5cm 2.0cm},clip]{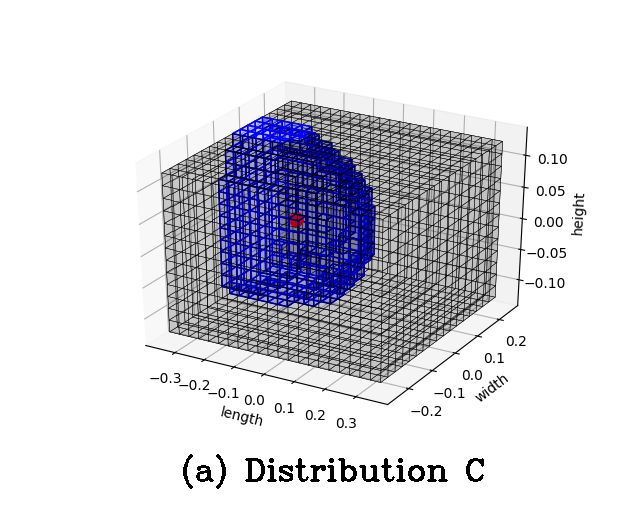}
  \end{minipage}
  \begin{minipage}[b]{0.2\textwidth}
    \centering
    \includegraphics[width=36mm,trim={2cm 1cm 1.5cm 2.0cm},clip]{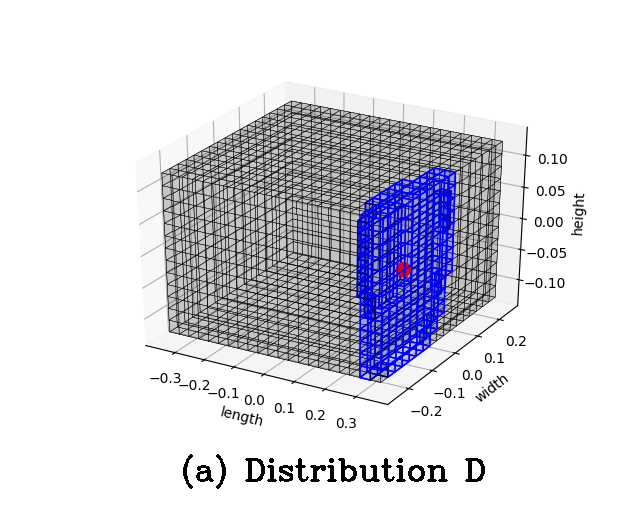}
  \end{minipage}
  \caption{\small{Mass distributions (referred to as Distributions A,B,C,D) used in the first set of experiments.}}
   \label{fig_distribs_exp_1}
\end{figure*}

  \section{Empirical results}
  \label{sec_experim}

We evaluate the proposed approach in a simulation (figure \ref{fig_simulation_description}(b)) developed in Pybullet \citep{coumans2021} based on rolling conveyor belts \citep{mcguire2009conveyors}. Varying mass distributions are generated using $3d$ gaussians approximated as voxel volumes, examples of which have been shown in figure \ref{fig_dynamics_and_example_distrib_pred} (c,d). We first compare the proposed approach to a classical baseline on four different distributions. Following the analysis of those results, we further validate our method on $30$ other distributions, in particular in terms of safe manipulation.

In order to reduce the complexity of the dynamics and in particular facilitate learning for the baseline, we restricted the actions by only allowing actions of the form $a=(a_1, a_2, 0, 0)$, $a=(0, 0, a_3, 0)$ or $a=(0, 0, 0, a_4)$. That is, we decoupled the application of velocity controls from position controls, and forbade the simultaneous adjustment of the positions $P_y^1, P_y^2$ of the conveyor belts. These restriction, which we imposed on experiments with both methods, had no incidence on the ability to control the box.

The mass distribution prediction network $\mathcal{H}$ that was used in all experiments was trained on short exploratory phases from $2000$ randomly generated boxes (see table \ref{table_training}). 

\subsection{Comparison to a classical baseline.}

We consider a classical MPC method using a black-box model of the form $\hat{s}_{t+1}=\mathcal{W}(s_t, a_t)$ with $\mathcal{W}$ a fully connected network of $\sim2.5e6$ parameters. First, the network is trained using data from several episodes until it reaches satisfactory accuracy. Then, the MPC control method, similar to the proposed approach, uses random shooting for generating actions and assigns $R_1, R_2$ scores to them based to the next predicted state. However, instead of using the equations of motion, it uses the trained $\mathcal{W}$.

Four different distributions, illustrated in figures \ref{fig_distribs_exp_1}(a,b,c,d), were used. We will refer to them by their assigned labels from the figure, \textit{i.e.} as distributions A, B, C, D. Note that each simulation instance is defined by the choice of box mass distribution. Distribution A was randomly generated, and its corresponding environment used to train $\mathcal{W}$. Distribution B was manually selected in order to be similar to A, so that we could expect that $\mathcal{W}$ would generalize reasonably well to this environment. Distribution C was also picked manually, to ensure that it was sufficiently different from A. Finally, distribution D, which is heavily skewed, was selected as an example of distributions that are \textit{uncontrollable} in a safe manner, as all of the mass will fall on one conveyor, resulting in important imbalance in terms of frictions. We \textit{do not} expect either method to succeed in rotating the box without the box becoming dangerously imbalanced during manipulation. However, the proposed method presents an advantage in this case: as it predicts the center of mass, detecting such potentially unsafe cases and aborting the manipulation is straight-forward.

We found that learning $\mathcal{W}$ to reach sufficient accuracy on distribution A was not possible using only data from random policies. Thus, in addition to those, we used data from trajectories obtained from already successful policies (from the physical priors pipeline). Details of the data used for training $\mathcal{W}$ are summarized in the second row of table \ref{table_training}. These results hint that the proposed approach is indeed less demanding in terms of exploration, and is more data-efficient.

Using the trained $\mathcal{H}$ and $\mathcal{W}$ networks, we executed each of the two pipelines $5$ times on each of the environments corresponding to the four distributions. The results are summarized in table \ref{table_comparison}. A notable difference is that $\mathcal{W}$ requires online adaptation, especially when the mass distribution is different from the one it has been trained on. We found that updating the model for one epoch (on all data from the current episode) every $20$ steps resulted in satisfying performance. The proposed approach, on the other hand, explicitly decouples the physical rules and relations that are common to all seen and unseen tasks from those that only hold in a particular environment. This allows it to be trained on multiple environments (as specified in table \ref{table_training}) and generalize in zero-shot manner, while classical black-box models such as $\mathcal{W}$ overfit a single environment. Indeed, one could improve the generalization of $\mathcal{W}$ through methods such as optimization based meta-learning \citep{hospedales2021meta}, however, this would exponentially increase the need in terms of data collection.

Both methods have a perfect success rate for distributions A and B. However, as expected from the previous paragraph, the success rate of the baseline drastically drops on distribution C, which unlike distribution B, is highly dissimilar to distribution A. 

 \begin{table*}
     \hspace*{0.7cm}
      \begin{tiny}
     \begin{tabular}{lccc}
       \hline
       \ \ \ \ \ & \pbox{10cm}{num necessary transitions} & \pbox{10cm}{data from}  & \pbox{10cm}{zero-shot generalization}\\
       \hline
       \pbox{10cm}{Model for physical priors estimation} & \pbox{10cm}{{$\sim$ 90k}} & {\pbox{10cm}{$2000$ different random boxes ($45$ fixed initial exploratory\\ steps for each) }}  & \pbox{10cm}{{Yes}}\\
       \hline
       \pbox{10cm}{Model from MPC (baseline)} & \pbox{10cm}{{$\sim$ 122k}} & {\pbox{10cm}{data from $\sim 200$ full episodes in a single environment, \\ using random exploration + known successful policies}}  & {\pbox{10cm}{No}}\\
       \hline
     \end{tabular}
      \end{tiny}
      \caption{\small{Summary of the data used for training the baseline and the model used to predict physical priors.}}
     \label{table_training}
   \end{table*}

   \begin{table*}
     \hspace*{0.7cm}
     \begin{tiny}
     \begin{tabular}{lccc}
       \hline
       \ \ \ \ \ & \pbox{10cm}{success ratio} & \pbox{10cm}{online adaptation} & \pbox{10cm}{average number of steps spent in environment} \\
       \hline
       distribution A, Physical priors based method &  100\% & \color{green}{None} &  \color{green}{238.8} \\
       \hline
       distribution A, Classical MPC (baseline) &  100\% & {1 epoch every $20$ steps} &  {367.4}  \\
       \Xhline{6\arrayrulewidth}
       distribution B, Physical priors based method &  100\% & \color{green}{None}   & 389.2 \\
       \hline
       distribution B, Classical MPC (baseline) &  100\% & {1 epoch every $20$ steps}  & \color{red}{269.6}  \\
       \hline
       \Xhline{6\arrayrulewidth}
       distribution C, Physical priors based method &  \color{green}{100\%} & \color{green}{None}  &  527.8  \\
       \hline
       distribution C, Classical MPC (baseline) &  {60\%} & {1 epoch every $20$ steps}   &  \color{red}{401.0} \\
       \Xhline{6\arrayrulewidth}
       distribution D, Physical priors based method & \multicolumn{2}{c}{\color{green}{Aborts execution as}} & \color{green}{\hspace*{-3.7cm}\ \  an unsafe distribution is detected} \\
       \hline
       distribution D, Classical MPC (baseline) &  {0\%} & {1 epoch every $20$ steps}  & N.A \\
     \end{tabular}
     \end{tiny}
     \caption{\small{Results of both methods on the four distributions of figure \ref{fig_distribs_exp_1}, averaged over five executions per distribution and method. Statistics for which the physics prior based pipeline yields better values are highlighted in green, and those were it was outperformed by the baseline are highlighted in red. We conjecture that the latter are mainly due to the uncertainty in the $a\mapsto \{F_1, F_2, F_3, F_4\}$ mapping. We note that contrary to the baseline, the proposed approach operates in a zero-shot manner in each environment, \textit{i.e.} it is not adapted to each new mass distribution.}}
     \label{table_comparison}
   \end{table*}

Adding prior information inferred from physics results in the prediction of meaningful physical quantities, and this interpretability, among other benefits, enables straight-forward implementation of sanity check and safety protocols. Distribution D clearly illustrates that: while MPC using $\mathcal{W}$ blindly tries to apply controls to this hazardous configuration, the prior-based method is able to abort execution as the predicted distribution is identified as unsafe.

Before closing this subsection, we note that the baseline outperforms the proposed approach in terms of numbers of timesteps required to solve the task. While we conjecture that this is due to the uncertainty from the $a\mapsto \{F_1, F_2, F_3, F_4\}$ mapping, we find it also important to emphasize that this lower number of timesteps does not translate to less time-consuming episodes: indeed, the time spend on the frequent on-line updates that are required by $\mathcal{W}$ outweighs any gains from the reduced number of timesteps.

\subsection{Validation on random distributions and safe manipulation}

We further validate the proposed approach, in particular in terms of safety, by evaluating its performance on $30$ different mass distributions. Two criteria are of importance in defining safety in the box rotation problems: 1) The evolution of the \textit{balance error} $err_{balance}=||r_{geom}-0.5(P_y^1+P_y^2)||_2^2$, which indicates how well $r_{geom}$ is centered between the two conveyors, and 2) whether the method is capable of aborting the manipulation early (\textit{i.e.} at the end of the exploration phase) for mass distributions that are physically hazardous to rotate. For the first criterion, based on the real robot, a hard threshold of $0.1\times box\_length$ is considered as the safety limit. The boxes used in our experiments are dimensions $40cms\times 30cms \times 15cms$, so this corresponds to $4.0cms$. For the second criterion, we must first define what we mean by hazardous. Noting the box as a volume $([0, l_{box}]\times [0, w_{box}]\times [0,h_{box}]$ with $l_{box}, w_{box}, h_{box} $ the length, width and height of the box, we partition the length and width in four equal parts: $\{[0, \frac{l_{box}}{4}], [\frac{l_{box}}{4}, \frac{l_{box}}{2}], [\frac{l_{box}}{2}, 3\frac{l_{box}}{4}], [3\frac{l_{box}}{4}, l_{box}]\}$ for the length and similarly for the width. We will say that a box mass distribution is \textit{hazardous} to rotate if its entire mass is concentrated in one of those volumes:

\begin{scriptsize}
\begin{equation}
  \begin{split}
    U_1 & \triangleq [0, \frac{l_{box}}{4}]\times[0,w_{box}]\times[0,h_{box}] \ \ \ \ \ \ \ \ \ \ \ \ \ \ \ \ \ \ \ U_2 \triangleq [3\frac{l_{box}}{4},l_{box}]\times[0,w_{box}]\times[0,h_{box}] \\
    U_3 & \triangleq [0, l_{box}]\times[0,\frac{w_{box}}{4}]\times[0,h_{box}] \ \ \ \ \ \ \ \ \ \ \ \ \ \ \ \ \ \ \ U_4 \triangleq [0, l_{box}]\times[3\frac{w_{box}}{4},w_{box}]\times[0,h_{box}].
  \end{split}
\end{equation}
\end{scriptsize}

For these experiments, we randomly sampled $30$ mass distributions that were \textit{non-hazardous}. On those environments, the proposed method yielded a success rate of $83.33\%$ (\textit{i.e.} $25$ successes out of $30$). Figure \ref{fig_mapping_and_balance}(c) summarizes the safety error $err_{balance}$ obtained with the proposed method. Two points are worth mentioning: first, one failure was due to an incorrect classification of one of the boxes as hazardous as a result of an inaccurate prediction $\hat{\Pi}$ for that environment. A second failure, also due to an inaccurate $\hat{\Pi}$ prediction, produces the maximum error shown as the dashed red curve in figure \ref{fig_mapping_and_balance}(c). The remaining three failure cases did not suffer from poor mass distribution predictions. We conjecture that failure in that case is caused by the uncertainties in the mapping $a\mapsto \{F_1, F_2, F_3, F_4\}$, which our method does not take into account. Note that the quality of the prediction $\hat{\Pi}$ was evaluated using the Intersection of Union between the prediction and the ground truth volume.

   \section{Limitations}

Training a network to estimate physical quantities can lead to additional supervision needs, which in turn results in difficulties for integration on real robots. In the context of the box rotation problem, training $\mathcal{H}$ requires knowledge of mass distributions. Therefore, learning based only on data from the real robot would necessitate the manufacture of specific mass distributions, \textit{e.g.} using 3d printed blocks. Our solution thus trades off efforts spent on setting up the experimental platform for data collection and training/computation time. A more balanced solution in terms of cost would be to combine the use of sim2real methods leveraging techniques such as domain randomization \citep{chaffre2020sim, nikdel2021lbgp, zhao2020sim} with limited data from the real robot. A second limitation of our method is that the physical priors were manually identified/derived. This requires more development efforts and can be error-prone. Thus, a direction that is worth exploring is the automatic extraction of priors, using \textit{e.g.} LLMs/foundation models \citep{bommasani2021opportunities}. Finally, we note that the solution we provided to the box rotation problem used a manually set fixed sequence of actions for exploration, which wouldn't be optimal for more complex problems. In such situations, learning dedicated exploration agents might be beneficial.
  
  \section{Conclusion}

   We presented a case study, based on the SOTO2 robot, that illustrates the advantages of incorporating physical priors in Model-based control. We replaced the black-box model of a sampling-based MPC  pipeline with a gray-box model that used a short exploration phase to estimate physically meaningful quantities. Our experiments showed increased data-efficiency, generalization and explainability, as well as more straight-forward implementation of safety requirements.

\clearpage
\acks{
  This work was supported by the European Union's H2020-EU.1.2.2 Research and Innovation Program through FET Project VeriDream (Grant Agreement No. 951992). We wish to thank Quentin Levent for their initial work on the conveyor belt simulation. We express our gratitude to Kathia Melbouci for their valuable input during the development of the presented material.}

\bibliography{example}  % .bib

\begin{thebibliography}{23}
\providecommand{\natexlab}[1]{#1}
\providecommand{\url}[1]{\texttt{#1}}
\expandafter\ifx\csname urlstyle\endcsname\relax
  \providecommand{\doi}[1]{doi: #1}\else
  \providecommand{\doi}{doi: \begingroup \urlstyle{rm}\Url}\fi

\bibitem[Bommasani et~al.(2021)Bommasani, Hudson, Adeli, Altman, Arora, von
  Arx, Bernstein, Bohg, Bosselut, Brunskill,
  et~al.]{bommasani2021opportunities}
Rishi Bommasani, Drew~A Hudson, Ehsan Adeli, Russ Altman, Simran Arora, Sydney
  von Arx, Michael~S Bernstein, Jeannette Bohg, Antoine Bosselut, Emma
  Brunskill, et~al.
\newblock On the opportunities and risks of foundation models.
\newblock \emph{arXiv preprint arXiv:2108.07258}, 2021.

\bibitem[Brunton et~al.(2016)Brunton, Proctor, and Kutz]{brunton2016sparse}
Steven~L Brunton, Joshua~L Proctor, and J~Nathan Kutz.
\newblock Sparse identification of nonlinear dynamics with control (sindyc).
\newblock \emph{IFAC-PapersOnLine}, 49\penalty0 (18):\penalty0 710--715, 2016.

\bibitem[Chaffre et~al.(2020)Chaffre, Moras, Chan-Hon-Tong, and
  Marzat]{chaffre2020sim}
Thomas Chaffre, Julien Moras, Adrien Chan-Hon-Tong, and Julien Marzat.
\newblock Sim-to-real transfer with incremental environment complexity for
  reinforcement learning of depth-based robot navigation.
\newblock \emph{arXiv preprint arXiv:2004.14684}, 2020.

\bibitem[Chalup et~al.(2007)Chalup, Murch, and Quinlan]{chalup2007machine}
Stephan~K Chalup, Craig~L Murch, and Michael~J Quinlan.
\newblock Machine learning with aibo robots in the four-legged league of
  robocup.
\newblock \emph{IEEE Transactions on Systems, Man, and Cybernetics, Part C
  (Applications and Reviews)}, 37\penalty0 (3):\penalty0 297--310, 2007.

\bibitem[Coumans and Bai(2016--2021)]{coumans2021}
Erwin Coumans and Yunfei Bai.
\newblock Pybullet, a python module for physics simulation for games, robotics
  and machine learning.
\newblock \url{http://pybullet.org}, 2016--2021.

\bibitem[Deisenroth and Rasmussen(2011)]{deisenroth2011pilco}
Marc Deisenroth and Carl~E Rasmussen.
\newblock Pilco: A model-based and data-efficient approach to policy search.
\newblock In \emph{Proceedings of the 28th International Conference on machine
  learning (ICML-11)}, pages 465--472, 2011.

\bibitem[Hospedales et~al.(2021)Hospedales, Antoniou, Micaelli, and
  Storkey]{hospedales2021meta}
Timothy Hospedales, Antreas Antoniou, Paul Micaelli, and Amos Storkey.
\newblock Meta-learning in neural networks: A survey.
\newblock \emph{IEEE transactions on pattern analysis and machine
  intelligence}, 44\penalty0 (9):\penalty0 5149--5169, 2021.

\bibitem[Islam et~al.(2021)Islam, Eberle, Ghafoor, and
  Ahmed]{islam2021explainable}
Sheikh~Rabiul Islam, William Eberle, Sheikh~Khaled Ghafoor, and Mohiuddin
  Ahmed.
\newblock Explainable artificial intelligence approaches: A survey.
\newblock \emph{arXiv preprint arXiv:2101.09429}, 2021.

\bibitem[Kaiser et~al.(2019)Kaiser, Babaeizadeh, Milos, Osinski, Campbell,
  Czechowski, Erhan, Finn, Kozakowski, Levine, et~al.]{kaiser2019model}
Lukasz Kaiser, Mohammad Babaeizadeh, Piotr Milos, Blazej Osinski, Roy~H
  Campbell, Konrad Czechowski, Dumitru Erhan, Chelsea Finn, Piotr Kozakowski,
  Sergey Levine, et~al.
\newblock Model-based reinforcement learning for atari.
\newblock \emph{arXiv preprint arXiv:1903.00374}, 2019.

\bibitem[Ljung(2017)]{wileybook}
Lennart Ljung.
\newblock \emph{System Identification}.
\newblock 2017.
\newblock \doi{https://doi.org/10.1002/047134608X.W1046.pub2}.
\newblock URL
  \url{https://onlinelibrary.wiley.com/doi/abs/10.1002/047134608X.W1046.pub2}.

\bibitem[Lutter et~al.(2021)Lutter, Silberbauer, Watson, and
  Peters]{lutter2021differentiable}
Michael Lutter, Johannes Silberbauer, Joe Watson, and Jan Peters.
\newblock Differentiable physics models for real-world offline model-based
  reinforcement learning.
\newblock In \emph{2021 IEEE International Conference on Robotics and
  Automation (ICRA)}, pages 4163--4170. IEEE, 2021.

\bibitem[McGuire(2009)]{mcguire2009conveyors}
Patrick~M McGuire.
\newblock \emph{Conveyors: application, selection, and integration}.
\newblock CRC Press, 2009.

\bibitem[Meriam et~al.(2020)Meriam, Kraige, and Bolton]{meriam2020engineering}
James~L Meriam, L~Glenn Kraige, and Jeff~N Bolton.
\newblock \emph{Engineering mechanics: dynamics}.
\newblock John Wiley \& Sons, 2020.

\bibitem[Moerland et~al.(2020)Moerland, Broekens, and
  Jonker]{moerland2020model}
Thomas~M Moerland, Joost Broekens, and Catholijn~M Jonker.
\newblock Model-based reinforcement learning: A survey.
\newblock \emph{arXiv preprint arXiv:2006.16712}, 2020.

\bibitem[Monta{\~n}o~Guzm{\'a}n(2003)]{montano2003fuzzy}
Linett Monta{\~n}o~Guzm{\'a}n.
\newblock Fuzzy measures and integrals.
\newblock \emph{Acta Nova}, 2\penalty0 (2):\penalty0 216--227, 2003.

\bibitem[Nagabandi et~al.(2018)Nagabandi, Kahn, Fearing, and
  Levine]{nagabandi2018neural}
Anusha Nagabandi, Gregory Kahn, Ronald~S Fearing, and Sergey Levine.
\newblock Neural network dynamics for model-based deep reinforcement learning
  with model-free fine-tuning.
\newblock In \emph{2018 IEEE International Conference on Robotics and
  Automation (ICRA)}, pages 7559--7566. IEEE, 2018.

\bibitem[Nikdel et~al.(2021)Nikdel, Vaughan, and Chen]{nikdel2021lbgp}
Payam Nikdel, Richard Vaughan, and Mo~Chen.
\newblock Lbgp: Learning based goal planning for autonomous following in front.
\newblock In \emph{2021 IEEE International Conference on Robotics and
  Automation (ICRA)}, pages 3140--3146. IEEE, 2021.

\bibitem[Pinneri et~al.(2021)Pinneri, Sawant, Blaes, Achterhold, Stueckler,
  Rolinek, and Martius]{pinneri2021sample}
Cristina Pinneri, Shambhuraj Sawant, Sebastian Blaes, Jan Achterhold, Joerg
  Stueckler, Michal Rolinek, and Georg Martius.
\newblock Sample-efficient cross-entropy method for real-time planning.
\newblock In \emph{Conference on Robot Learning}, pages 1049--1065. PMLR, 2021.

\bibitem[Salehi et~al.(2023)Salehi, Rühl, and Doncieux]{salehi2022meta}
Achkan Salehi, Steffen Rühl, and Stephane Doncieux.
\newblock Adaptive asynchronous control using meta-learned neural ordinary
  differential equations.
\newblock \emph{IEEE Transactions on Robotics}, 2023.
\newblock \doi{10.1109/TRO.2023.3326350}.

\bibitem[Sanyal and Roy(2022)]{sanyal2022ramp}
Sourav Sanyal and Kaushik Roy.
\newblock Ramp-net: A robust adaptive mpc for quadrotors via physics-informed
  neural network.
\newblock \emph{arXiv preprint arXiv:2209.09025}, 2022.

\bibitem[Sekar et~al.(2020)Sekar, Rybkin, Daniilidis, Abbeel, Hafner, and
  Pathak]{sekar2020planning}
Ramanan Sekar, Oleh Rybkin, Kostas Daniilidis, Pieter Abbeel, Danijar Hafner,
  and Deepak Pathak.
\newblock Planning to explore via self-supervised world models.
\newblock In \emph{International Conference on Machine Learning}, pages
  8583--8592. PMLR, 2020.

\bibitem[Seo et~al.(2021)Seo, Chen, Shin, Lee, Abbeel, and Lee]{seo2021state}
Younggyo Seo, Lili Chen, Jinwoo Shin, Honglak Lee, Pieter Abbeel, and Kimin
  Lee.
\newblock State entropy maximization with random encoders for efficient
  exploration.
\newblock In \emph{International Conference on Machine Learning}, pages
  9443--9454. PMLR, 2021.

\bibitem[Zhao et~al.(2020)Zhao, Queralta, and Westerlund]{zhao2020sim}
Wenshuai Zhao, Jorge~Pe{\~n}a Queralta, and Tomi Westerlund.
\newblock Sim-to-real transfer in deep reinforcement learning for robotics: a
  survey.
\newblock In \emph{2020 IEEE Symposium Series on Computational Intelligence
  (SSCI)}, pages 737--744. IEEE, 2020.

\end{thebibliography}
\end{small}
\end{document}